\documentclass{article}

\PassOptionsToPackage{numbers}{natbib}
\usepackage[final]{neurips_2022}

\usepackage[utf8]{inputenc} %
\usepackage[T1]{fontenc}    %
\usepackage{hyperref}       %
\usepackage{booktabs}       %
\usepackage{amsfonts}       %
\usepackage{nicefrac}       %
\usepackage{microtype}      %
\usepackage{xcolor}         %

\title{Lempel-Ziv Networks}

\author{%
  Rebecca Saul \\ 
  Booz Allen Hamilton \\
  Laboratory for Physical Sciences\\
  \texttt{Saul\_Rebecca@bah.com} \\ 
   \And 
  Mohammad Mahmudul Alam \\
  University of Maryland, Baltimore County\\
  \texttt{m256@umbc.edu} \\
   \AND
  John Hurwitz \\ 
   Laboratory for Physical Sciences \\ 
   University of Maryland, Baltimore County \\ 
   \And 
  Edward Raff \\
   Booz Allen Hamilton \\
  Laboratory for Physical Sciences \\
  University of Maryland, Baltimore County \\
   \texttt{Raff\_Edward@bah.com} \\
   \And
  Tim Oates \\
  University of Maryland, Baltimore County\\
   \texttt{oates@umbc.edu} \\
   \And 
  James Holt \\
  Laboratory for Physical Sciences\\
   \texttt{holt@lps.umd.edu} \\
}

\usepackage[utf8]{inputenc} %
\DeclareUnicodeCharacter{2212}{-}

\usepackage{microtype}
\usepackage{graphicx}
\usepackage{booktabs} %
\usepackage{multirow}
\usepackage{siunitx}
\usepackage{todonotes}
\usepackage{wrapfig}
\usepackage{afterpage}
\usepackage{float}
\usepackage{sidecap}
\usepackage{listings}
\usepackage[gen]{eurosym} %
\usepackage{fancyhdr}
\usepackage{lastpage}
\usepackage{tikz}
\usepackage{mathdots}
\usepackage{yhmath}
\usepackage{cancel}
\usepackage{color}
\usepackage{siunitx}
\usepackage{array}
\usepackage{multirow}
\usepackage{amssymb}
\usepackage{gensymb}
\usepackage{tabularx}
\usepackage{booktabs}
\usetikzlibrary{fadings}
\usetikzlibrary{patterns}
\usetikzlibrary{shadows.blur}
\usepackage{enumitem}
\usepackage{subfigure}
\usepackage[utf8]{inputenc}
\usepackage{pgfplots}
\usepackage{placeins}
\usepackage{natbib}
\setitemize{noitemsep,topsep=0pt,parsep=0pt,partopsep=0pt}

\usepackage{caption}
\usepackage{appendix}
\usepackage{amsmath,bm}
\usepackage{amssymb}
\usepackage[autostyle, english=american]{csquotes}
\MakeOuterQuote{"}

\usepackage{adjustbox}

\newcommand\textcite{\citet}
\usetikzlibrary{shapes.geometric, arrows, fit, backgrounds}
\tikzstyle{net} = [regular polygon, regular polygon sides=6, text width=1.5cm, minimum height = 1.5cm, minimum width=1.5cm, text centered, draw=black, fill=orange!30]
\tikzstyle{func} = [circle, minimum width=2cm,  text centered, draw=black, fill=teal!30]
\tikzstyle{arrow} = [thick,->,>=stealth]

\DeclareUnicodeCharacter{2212}{−}
\usepgfplotslibrary{groupplots,dateplot}
\usetikzlibrary{patterns,shapes.arrows}
\pgfplotsset{compat=newest}
\newcommand{\bind}{%
  \mathrel{%
\begin{tikzpicture}[x=0.75pt,y=0.75pt,yscale=-1,xscale=1]
\draw   (0,5) .. controls (0,2.24) and (2.24,0) .. (5,0) .. controls (7.76,0) and (10,2.24) .. (10,5) .. controls (10,7.76) and (7.76,10) .. (5,10) .. controls (2.24,10) and (0,7.76) .. (0,5) -- cycle ;
\draw    (5,0) -- (1.57,8.57) ;
\draw    (8.47,1.43) -- (5,10) ;
\draw    (0.33,3.33) -- (9.67,3.33) ;
\draw    (0.33,6.67) -- (9.67,6.67) ;
\end{tikzpicture}
  }%
}

\begin{document}

\maketitle

\begin{abstract}
Sequence processing has long been a central area of machine learning research. Recurrent neural nets have been successful in processing sequences for a number of tasks; however, they are known to be both ineffective and computationally expensive when applied to very long sequences. Compression-based methods have demonstrated more robustness when processing such sequences --- in particular, an approach pairing the Lempel-Ziv Jaccard Distance (LZJD) with the k-Nearest Neighbor algorithm has  shown promise on long sequence problems (up to $T=200,000,000$ steps) involving malware classification. Unfortunately, use of LZJD is limited to discrete domains. To extend the benefits of LZJD to a continuous domain, we investigate the effectiveness of a deep-learning analog of the algorithm, the Lempel-Ziv Network. While we achieve successful proof of concept, we are unable to improve meaningfully on the performance of a standard LSTM across a variety of datasets and sequence processing tasks. In addition to presenting this negative result, our work highlights the problem of sub-par baseline tuning in newer research areas.  
\end{abstract}

\maketitle

\section{Introduction} \label{sec:intro}
Sequence processing has been an important focus of the machine learning community for decades. Due to the ubiquity of sequential data, there is a need for effective algorithms to process this data for a diverse set of tasks, including classification and time series analysis. Recurrent architectures such as long short-term memory (LSTM)~\cite{Hochreiter1997} have been successful in processing sequences for a number of such tasks in recent years \cite{yu2019lstms}. A known issue of recurrent networks, however, is that they tend to be both ineffective \cite{trinh2018learning} and computationally expensive \cite{lipton2015critical} when applied to very long sequences.These challenges have hindered efforts to learn long-term dependencies over sequential data with recurrent neural nets (RNNs). 
\par
Compression-based methods have demonstrated more robustness in tackling the issues associated with long sequences. The Lempel-Ziv Jaccard Distance (LZJD) ~\cite{raff_lzjd_2017,Raff_Nicholas_2018,Raff_Aurelio_Nicholas_2019,Raff_Nicholas_2017} takes the Lempel-Ziv compression of sequences before applying the Jaccard distance to the compressions. An approach pairing LZJD with the k-Nearest Neighbor algorithm has been shown to be both more accurate and orders of magnitude faster than previous techniques for malware classification from byte sequences. However, LZJD is limited by the fact that it is only applicable in a discrete domain; moreover LZJD is unable to learn about or exploit the existence of similar compressions since it recognizes only exact matches. This is not ideal for use in areas such as malware classification, where the underlying data is continually evolving. 
\par 
We explore the effectiveness of a deep-learning analog of LZJD, the Lempel-Ziv Network, at extending LZJD's benefits to the continuous domain. Despite establishing successful proof of concept, we fail to consistently improve on the performance of a traditional LSTM over a collection of datasets and sequence processing tasks.
\par 
The rest of the paper is organized as follows. In Section \ref{sec:background}, we review LZ compression and various types of associative memories. In Section \ref{sec:lzlayer}, the novel Lempel-Ziv based recurrent layer is delineated. Experimental details and results are presented in Section \ref{sec:results}. Finally, we conclude in Section \ref{sec:conclusion}. 

\section{Background} \label{sec:background}
\paragraph {LZ compression}
\begin{wraptable}[2]{r}{0.31\textwidth}
\vspace{-55pt}
    \centering
    \caption{LZ compression on two sequences.}
    \vspace{-5pt}
        \begin{tabular}{|c|c|}
        \hline
         Sequence    &  Compression \\
         \hline
         aabbaba    & \{a, ab, b, aba\} \\
         \hline
         aabbba & \{a, ab, b, ba\} \\
         \hline
        \end{tabular}
    \label{tab:LZCompression}
\end{wraptable}
Lempel-Ziv Jaccard Distance (LZJD)~\cite{raff_lzjd_2017} uses Lempel-Ziv (LZ) compression to condense long sequences into a compact representation consisting of a set of unique subsequences. LZ compression works by keeping an external memory that store unique subsequences of the input sequence. To determine which subsequences are placed into the external memory, we traverse the input sequence with a sliding window. If the subsequence currently selected by the sliding window is already stored in the external memory, then we increment the "end" index of the window and keep the "start" index fixed, increasing the size of our window until the subsequence it identifies is not present in the external memory. Once such a sequence is found, we insert it into the external memory and move the "start" index of our sliding window to the current "end" index. We repeat this process to locate and insert successive unique subsequences into the external memory until we have covered the length of the input sequence.

\par
This explication of LZ compression also reveals some of its weaknesses. Let $A, B, C$ be subsequences present in an external memory arising from LZ compression of a sequence. Let $A = \{1.29, 7.89, 0.11\}$, $B = \{1.28, 7.91, 0.10\}$, and $C = \{5.01, 2.63, 2.17\}$. It is clear that $A$ and $B$ are very similar to each other, and both are substantially different from $C$. Yet the LZ compression algorithm is unable to express this nuance and treats all three subsequences as being equally dissimilar, and so we end up with two nearly identical subsequences, $A$ and $B$, in our external memory. We hope that by combining the insights of LZJD with the functionality of an RNN, we can give LZ compression the flexibility to handle similar subsequences while maintaining its effectiveness as a tool for finding compact representations of long sequences.

\paragraph {Associative memories}
We require an associative memory to perform two essential functions - we must be able to insert new items into the memory, and we must be able to query the memory for information about items it already contains. This enables an associative memory to mimic the function of LZJD's external memory. By first querying the memory for information about prospective inputs, we can be sure to input only items (i.e. subsequences or their representations) that have not been previously inserted. Here we review the three types of associative memories with which we experimented in our LZ Layer.

\textit{Hopfield Networks} \cite{hopfield1982neural} are among the best known associative memories. Designed to store and retrieve patterns, in response to a query $q$ Hopfield networks will return the pattern in memory most similar to $q$, or an average of similar patterns. In \cite{ramsauer2020hopfield}, Ramsauer et al. develop a modern Hopfield network for continuous states, with exponential storage capacity. Because patterns can be retrieved from memory in one update step, this Hopfield network is ideal for use in deep learning architectures. We utilize a static version of their \textit{Hopfield} layer to perform the insert and query functions of associative memory.

\textit{Vector Symbolic Architectures} (VSA) use high-dimensional vectors to imitate symbolic processing. VSAs are typically equipped with three crucial operations - a bundling operation, a binding operation, and an unbinding operation \cite{schlegel2020comparison}. The bundling operation takes in two input vectors and outputs a third vector similar to both inputs, and is denoted by a simple addition sign. The binding operation $\mathfrak B$ is used to pair two vectors together, while the unbinding operation $\mathfrak B^+$ undoes this pairing. For example, by assigning concepts such as ``Age'' and values such as ``30'' to high-dimensional vectors, we could set $\textrm{John} = \mathfrak B(\textrm{Age}, 30).$ Then, using the unbinding function, we could perform a lookup, resulting in $\mathfrak B^+(\textrm{John}, \textrm{Age}) = 30.$ By combining binding and bundling, we can create more complex representations, such as $\textrm{Jane} = \mathfrak B(\textrm{Age}, 25) + \mathfrak B(\textrm{Weight}, 120).$ However, the accuracy of the unbinding operation decreases as the number of terms bundled together increases. Maximizing this accuracy as bundle size increases is a key consideration in the choice of binding operation.
\par 
With a few additional specifications, a VSA can act as an associative memory. First, we fix a ``tag'' vector. Then, we bind every vector we want to insert into the associative memory to the tag vector. Our memory then becomes the bundling of all the (vector, tag) pairs. To add a new vector $v$ to the memory, we simply bundle again: $\textrm{memory}_{new} = \textrm{memory}_{old} + \mathfrak B(v, \textrm{tag}).$ To query the memory for a vector $q$, we compute $\mathfrak B^+(\textrm{memory}, q) = \widehat{\textrm{tag}}.$ If $\widehat{\textrm{tag}} \approx \textrm{tag}$, we conclude that $q$ is stored in the associative memory. On the contrary, if $\widehat{\textrm{tag}} \not \approx \textrm{tag}$, we can assume $q$ is not in the associative memory.
\par 
In this paper, we construct associative memories using a VSA. For the bundling operation, we use simple element-wise addition. For the binding and unbinding operations, we experimented with the two different paradigms discussed in the remainder of this section. 

\textit{Vector-Derived Transformation Binding} (VTB) defines the binding operation $\mathfrak{B}$ as follows: for $d$ a perfect square, $d' = d^{1/2}$, we have $\mathfrak B:\mathbb R^d \times \mathbb R^d \rightarrow \mathbb R^d$ where
$
\mathfrak B(x, y) = V_yx = 
\begin{bmatrix}
V_y' & 0  \\
0 & \ddots \end{bmatrix}x
$
and \autoref{eq:vtb}.

\begin{wrapfigure}[6]{r}{0.5\columnwidth}
\vspace{-20pt}
\begin{equation} \label{eq:vtb}
V_y' = d^{1/4}
\begin{bmatrix} y_1 & y_2 & \cdots & y_{d'} \\
y_{1+ d'} & y_{2 + d'} & \cdots & y_{2d'}  \\
\vdots & \vdots & \ddots & \vdots \\
y_{1+ (d-d')} & y_{2+ (d-d')} & \cdots & y_d \end{bmatrix}
\end{equation}
\end{wrapfigure}

The unbinding operation, the approximate inverse of $\mathfrak B$, is specified by $\mathfrak B^+(x, y)  = V_y^\top x$. In
\cite{gosmann2019vector}, Gosmann and Eliasmith found VTB to be an improvement over traditional binding operations such as circular convolution, as the associated unbinding operation retained a higher accuracy even when many terms were bound together. Additionally, with VTB, increased bindings had less effect on the vector norm, making VTB more suited for use in neural networks.

\textit{Holographic Reduced Representation} (HRR) provides an alternative way of implementing symbolic AI. Our HRR associative memory works very similarly to its VTB counterpart - only the binding operation is changed. 
\par 
HRR was first introduced by in \cite{plate1995holographic}. Circular convolution and fast Fourier transforms were used to define a binding operation $\bind$, where $s = a \bind b = \mathcal F^{-1}(\mathcal F(a) \odot \mathcal F(b)).$ By defining an identity function $\mathcal F(a^+)\mathcal F(a) = 1$, an inverse $a^+$ for $a$ is established, where $a^+~=~\mathcal F^{-1}\Big(\frac{1}{\mathcal F(a)}\Big).$ Then, unbinding proceeds as follows: $s \bind a^+ \approx b.$
\par
This HRR, however, has some considerable limitations. First, the inverse $a^+$ is numerically unstable, necessitating the use of a pseudo-inverse, differing by the reciprocal of the complex magnitude, in its place. Additionally, once more than ten bound pairs are bundled together, the accuracy of the unbinding operation, used for querying in the context of associative memories, is severely diminished. These concerns are mitigated in work done by~\cite{hrrxml,Alam2022}; \cite{hrrxml} defines a new complex unit magnitude projection which, when applied to each input of the binding operation before binding takes place, allows for accurate unbinding with up to 100$\times$ more bound vectors. Furthermore, by ensuring that all the vectors used in the HRR are unitary, the projection guarantees that the true inverse is equivalent to the numerically stable and computationally cheaper pseudo-inverse, eliminating one source of error and providing major speed-ups during implementation. We make use of this improved HRR in building our HRR Associative Memory.

\section{The Lempel-Ziv Layer} \label{sec:lzlayer}

\begin{wrapfigure}[5]{r}{0.7\textwidth}
\vspace{-50pt}
\centering
\resizebox{.7\columnwidth}{!}
{
\begin{tikzpicture} [node distance=2.4cm]

\node (rnn) [net] {RNN Cell};
\node (query) [right of =rnn, above of=rnn, circle, draw=black, fill=teal!30, xshift=-0.2cm] {Query};
\node (r) [right of = query, xshift=-1.3cm, draw=magenta] {r};
\node (insert) [right of = r, circle, draw=black, fill=teal!30, xshift=-1.3cm] {Insert};
\node (mem) [above of=query, text width=2cm, text centered, xshift=1.3cm, yshift=-1cm] {Associative Memory};
\node (nov) [net, right of=insert, above of=insert, xshift=0.4cm] {Novelty Score};
\node (reset) [func, right of=nov, below of=nov] {Reset};
\node (rnn2) [net, right of=reset, below of=reset, xshift=1cm] {RNN Cell};
\begin{scope} [on background layer]
\node [draw=black, fit={(query) (mem) (insert)}, fill=yellow!30] {};
\end{scope}

\draw [arrow] (rnn) -| node[anchor=south west, pos=0.02mm] {$\hat{h}_t$} (query);
\draw [arrow] (rnn) -| (insert);
\draw [arrow] (rnn) -| (reset);

\draw [arrow] (query) |- node[anchor=south west] {$\hat{r}$}([yshift=1cm]nov);
\draw [arrow] (r) |- +(1.4,0.8) |- ([yshift=-0.6cm]nov);
\draw [arrow] (nov) |- (insert);
\draw [arrow] (nov) |- node[anchor=south west, pos=0.1mm] {$p$}(reset);
\draw [arrow] (reset) -- node[anchor= south west, pos=0.002mm] {$h_t$} +(1.5,0) |- (rnn2);

\coordinate[left of=rnn] (hidden1);
\coordinate[right of=rnn2] (hidden2);
\coordinate[below of=rnn] (hidden3);
\coordinate[below of=rnn2] (hidden4);

\draw [arrow] (hidden1) -- node[anchor=south] {$h_{t-1}$} (rnn);
\draw [arrow] (rnn2) --  node[anchor=south] {$\hat{h}_{t+1}$} (hidden2);
\draw [arrow] (hidden3) -- node[anchor=west] {$x_t$} (rnn);
\draw [arrow] (hidden4) --  node[anchor=west] {$x_{t+1}$} (rnn2);

\end{tikzpicture}
}
\caption{Diagram of a Lempel-Ziv layer.}
\label{fig: LZDiagram}
\end{wrapfigure}
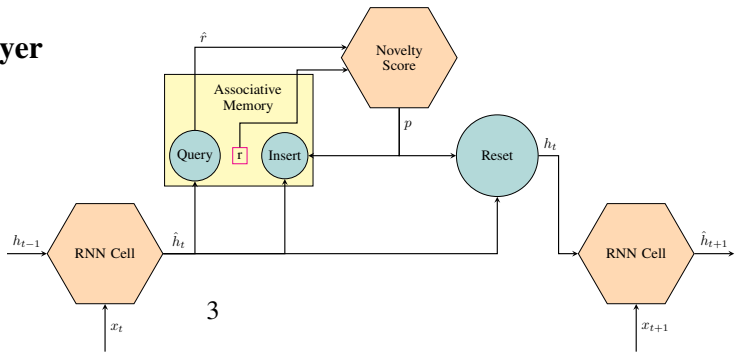

The LZ layer (see Figure \ref{fig: LZDiagram}) begins with a standard RNN Cell, which takes in the hidden state from the previous time step ($h_{t-1}$) and the input for the current time step ($x_t$) and outputs a new preliminary hidden state ($\hat{h}_t$). In a traditional RNN, we might stop here, and just consider $\hat{h}_t$ to be $h_t$, the hidden state which is passed along as input when the next item in the sequence is processed by the RNN. However, in the LZ layer, we want to perform the LZ digest-creating behavior on this preliminary hidden state $\hat{h}_t$, as the hidden state is the neural representation of the subsequence of data that the traditional LZ compression algorithm would analyze. 
\par 
Thus, we query the associative memory to ask if we have seen $\hat{h}_t$ before. The query returns a reconstructed vector $\hat{r}$, which we compare to a vector $r$ --- the tag vector if using a VTB or HRR associative memory, or $\hat{h}_t$ itself if using a Hopfield associative memory. The closer $\hat{r}$ is to $r$, the more similar $\hat{h}_t$ is to something already stored in memory. We pass $\hat{r}$ and $r$ into a learnable bilinear layer, combined with a sigmoid activation function and a Bernoulli layer, to determine if $\hat{h}_t$ is sufficiently dissimilar to anything we have seen before, i.e., if $\hat{r}$ is sufficiently far from $r$, to insert $\hat{h}_t$ into the associative memory as a new value. We refer to this bilinear-sigmoid-Bernoulli combination as a novelty score, and it returns $p=1$ if it judges $\hat{h}_t$ to be new enough to insert, and $p=0$ otherwise.~\footnote{By removing the final Bernoulli layer, we can make the novelty score $p$ a continuous value between 0 and 1, which we can interpret as the probability that $\hat{h}_t$ has not previously been inserted into the associative memory.} Because the novelty score is learnable, we theorize that it will allow our LZ Layer to treat similar, though not identical, subsequence representations differently than it treats vastly different subsequences, avoiding the problem with traditional LZ compression described in Section \ref{sec:background}. Once $p$ is computed, $p * \hat{h}_t$ is inserted into the associative memory. This is the neural equivalent of inserting unseen subsequences into the external memory during LZ compression.

\par
As the last step of the neural layer, we set $h_t~=~(1~-~p)~*~\hat{h}_t$. Setting $h_t$ determines the length of the subsequence to consider next for insertion into the associative memory, which in traditional LZ compression is done by adjusting the boundaries of the sliding window that passes over the input sequence. If we have just inserted $\hat{h}_t$ into the associative memory ($p=1$), that is an indication that we have identified a unique subsequence of the input. Thus, when looking for the next unique subsequence, we do not wish to consider any of the previous data points in the time series. In LZ compression, we set the starting index of the sliding window to the current endpoint. In the neural model, this corresponds to clearing out the hidden state, and setting $h_t=0$, effectively erasing any knowledge of prior steps in the sequence. However, if we have not inserted $\hat{h}_t$ (i.e. $p=0$), that means our current subsequence is not unique, so we would like to extend the subsequence by one time-step and evaluate its novelty again. In that case, we do not reset $h_t$, and let $h_t = \hat{h}_t$. Once we determine $h_t$, we are ready to process the next step in the time series. After processing the entire sequence, we can return a list of all the $\hat{h}_t$'s, with a corresponding mask of $p$’s indicating which $\hat{h}_t$'s were significant enough to input into the associative memory. We can also return the associative memory itself, which will give a different representation of this same information.

\section{Experiments and results} \label{sec:results}

\paragraph{Addition problem}
\begin{wraptable}[2]{o}{0.47\textwidth}
\vspace{-50pt}
\centering
\caption{Sample input for the addition problem: the desired output is 0.56+0.49=1.05.}
\begin{tabular}{|c|c|c|c|c|c|}
\hline
0.33 & 0.56 & 0.78 & 0.21 & 0.49 & 0.83 \\
\hline
0 & 1 & 0 & 0 & 1 & 0 \\
\hline
\end{tabular}
\label{tab: addprob}
\end{wraptable}

We began by testing our LZ layer, implemented in PyTorch, on the addition problem, a classic example used to assess RNN performance. The input to the addition problem is two sequences of equal length. The first sequence contains real numbers between zero and one, uniformly sampled. The second sequence, considered an indicator sequence, is all zeroes except for two randomly chosen entries, one in the first half of the sequence and one in the second half of the sequence, which are ones. The desired output of the RNN is the sum of the two entries in the first sequence that are selected by the indicator sequence (see Table \ref{tab: addprob}). RNNs are evaluated against a baseline model which predicts a sum of 1, regardless of the input sequence; this model has an expected mean squared error (MSE) of 0.167. Our setup of the addition problem mirrored as best as possible that of \cite{rotman2020shuffling}.

\par
\begin{wrapfigure}[12]{l}{0.50\textwidth}
\vspace{-15pt}
\centering
\includegraphics[width=0.5\textwidth]{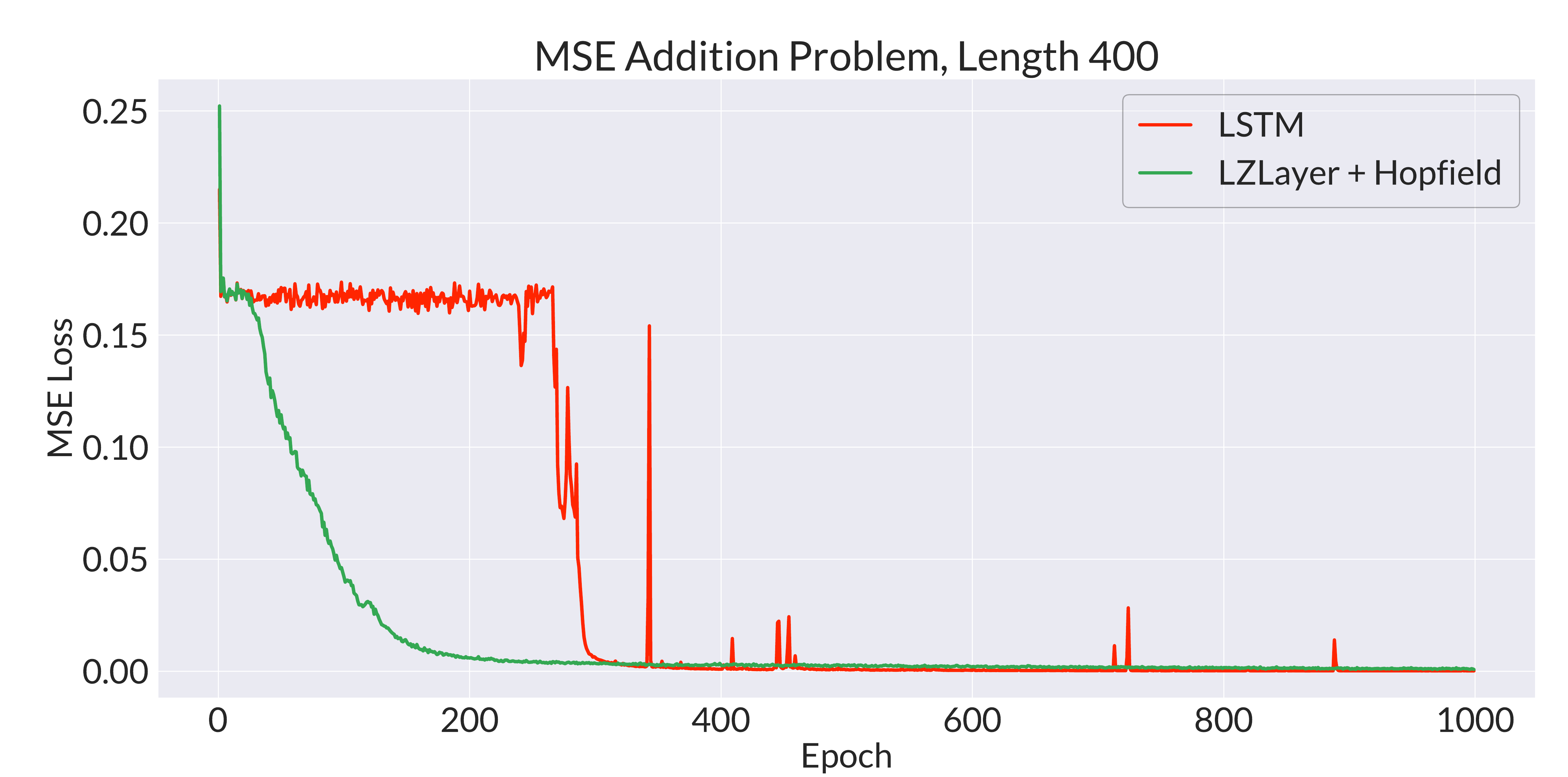}
\caption{Addition problem results --- LZ Layer with Hopfield Associative Memory.}
\label{fig: lzhopresults}
\end{wrapfigure}

First, we present results from an LZ Layer with a Hopfield Network Associative Memory and LSTM RNN cell (LZHOP). For a sequence length of 200, we set a hidden size of 128 and a batch size of 256. For a sequence length of 400 we keep the same hidden size, but decrease the batch size to 50 due to memory constraints. For all experiments, we use the RMSProp optimizer with a learning rate of $10^{-3}$ and a decay rate of 0.9, and our model contains only one LZ Layer. We include results from a traditional LSTM, trained with the same parameters, as a point of comparison. In Figure \ref{fig: lzhopresults}, we see that the LZHOP Layer reaches final error rates comparable to those of an LSTM, and can be trained in fewer epochs. This speed advantage over an LSTM grows as the sequence length increases (see Appendix \ref{sec:appendix}).
\par
\begin{figure}[H]
\vspace{-13pt}
\centering
\begin{minipage}{.5\textwidth}
\centering
\captionsetup{width=.9\linewidth}
\includegraphics[width=.99\linewidth]{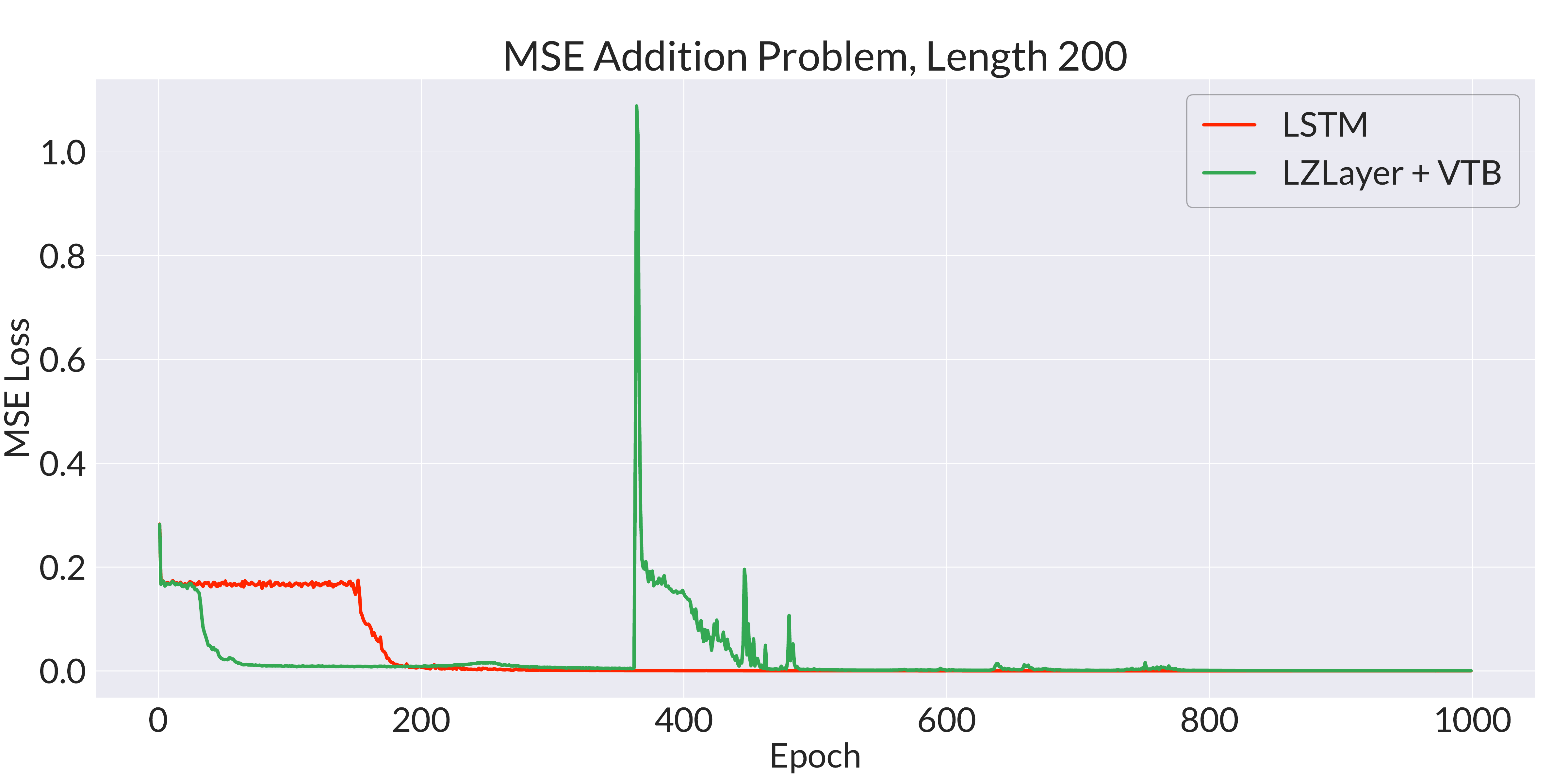}
\captionof{figure}{Addition problem results --- LZ Layer with VTB Associative Memory}
\label{fig: lzvtbresults}
\end{minipage}%
\begin{minipage}{.5\textwidth}
\centering
\captionsetup{width=.9\linewidth}
\includegraphics[width=.99\linewidth]{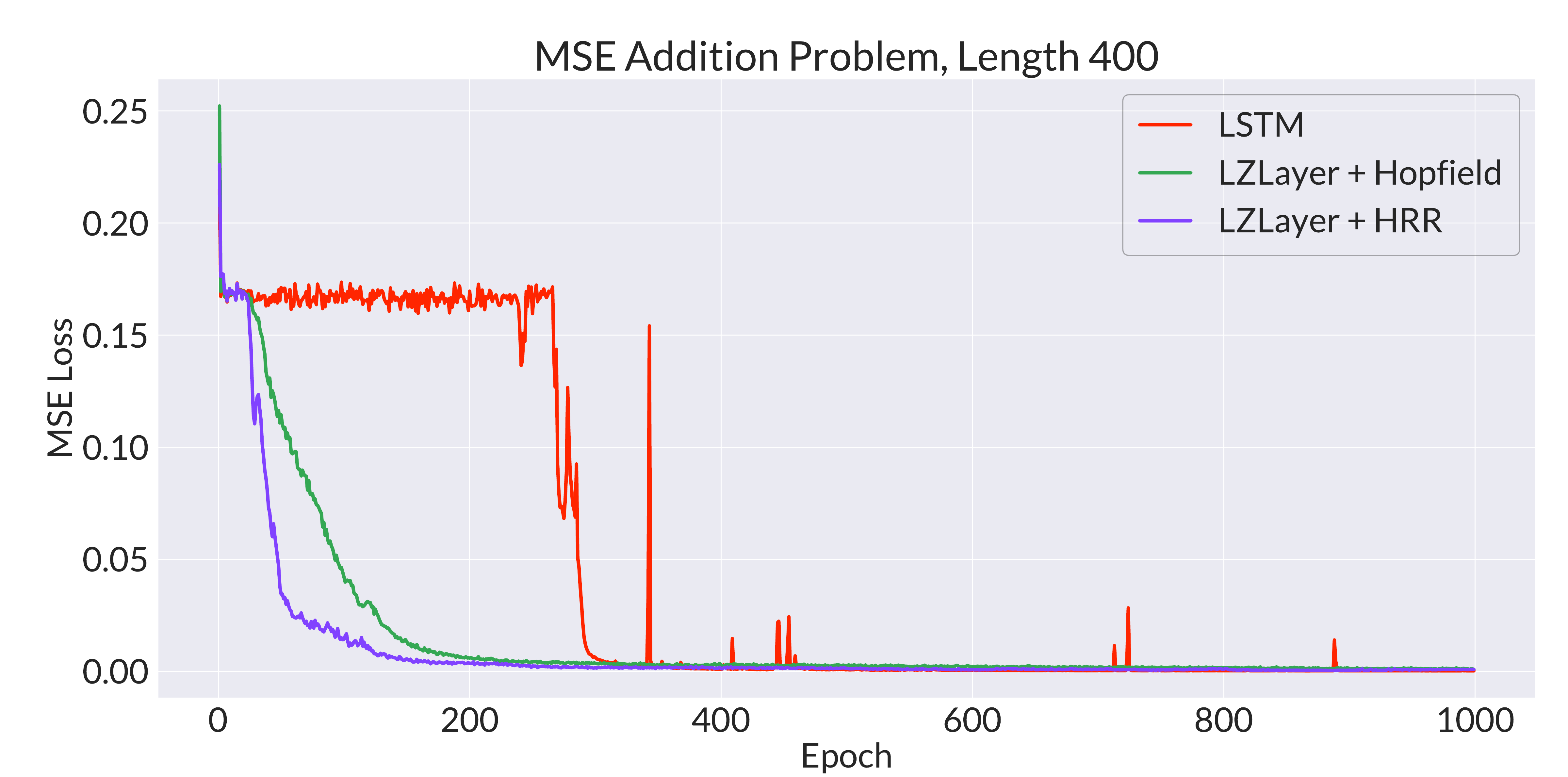}
\captionof{figure}{Addition problem results --- LZ Layer with HRR Associative Memory}
\label{fig: lzhrrresults}
\end{minipage}
\end{figure}

Although the initial results from the LZHOP Layer were very promising, this model was not scalable to longer sequence lengths because the Hopfield Network associative memory had quadratic run-time behavior. We next turned our attention to associative models requiring less computational overhead, with the hope that they would replicate the improvements achieved in the LZHOP Layer. We began with the LZ Layer with a VTB Associative Memory and LSTM RNN cell (LZVTB). In each experiment, we use a batch size of 50, with a hidden size of 256, as the VTB memory requires that the hidden size be a perfect square. Again, we use an RMSProp optimizer with a learning rate of $10^{-3}$ and a decay rate of 0.9 and compare our results to a traditional LSTM with the same parameters. In Figure \ref{fig: lzvtbresults}, we see that the LZVTB Layer also reaches error rates comparable to those of an LSTM, and is still able to train in fewer epochs. However, there are occasional dramatic spikes in the loss, indicating undesirable unstable behavior in the model. We believe this behavior occurred because not every vector inserted into the VTB Associative Memory was orthogonal.

\vspace{-15pt} To address some of the stability issues with the VTB Associative Memory, we turned to a new, HRR-based associative memory. We used the same parameters as in the LZHOP experiments, to enable easy comparison of the results, and found that an LZ Layer with an HRR Associative Memory and LSTM RNN cell (LZHRR) was able to replicate the stability of the LZHOP model while being less computationally expensive and requiring less memory. In addition to approaching comparable error rates, the LZHRR model also required fewer epochs to train than the LZHOP model (see Figure \ref{fig: lzhrrresults}). This evidence persuaded us that the LZHRR model would be a viable approach to many sequence processing problems.

\paragraph{UCR Time Series Archive}
The UCR Time Series Archive \cite{UCRArchive2018} is a collection of 128 datasets for time series classification problems created to ameliorate the phenomenon of cherry-picking datasets that yield good results in ML research. We trained an LZHRR model, along with an LSTM baseline model, on every single dataset in the UCR Archive, and the full results are shown in Table \ref{tab:ucr} in Appendix \ref{sec:appendix}. For both the LZ Layer and the LSTM, we chose a hidden size of 256. The models were trained using the Adam optimizer with a learning rate of $10^{-3}$ for 500 epochs. For the LZ Layer, we additionally tested the performance with various bias $b$ initializations over the Bilinear layer, where $b \in \{0, 1, -1\}$. Figure \ref{fig:improve_bias} shows the differences in accuracy between the LZ Layer and the LSTM on the UCR datasets. Overall, the LZ Layer outperforms a traditional LSTM in $46.88\%$ of the datasets, but there are instances in which it significantly underperforms the LSTM. Moreover, to run the UCR trials, we implemented our LZ Layer and LSTM in both PyTorch and JAX, the deep learning library from Google. Implementation in JAX dramatically improved the performance of both models in terms of computational time and accuracy. The accuracy improvement is shown in Figure \ref{fig:improve_jax} in Appendix \ref{sec:appendix} --- LSTM accuracy is improved on 69.53\% of datasets, and LZ Layer accuracy is improved on 79.68\% of datasets. 
\afterpage{
\begin{figure*}[!htbp]
\centering 
\begin{minipage}{0.5\textwidth}
\subfigure[Bias: 0]
{\includegraphics[width=0.99\textwidth]{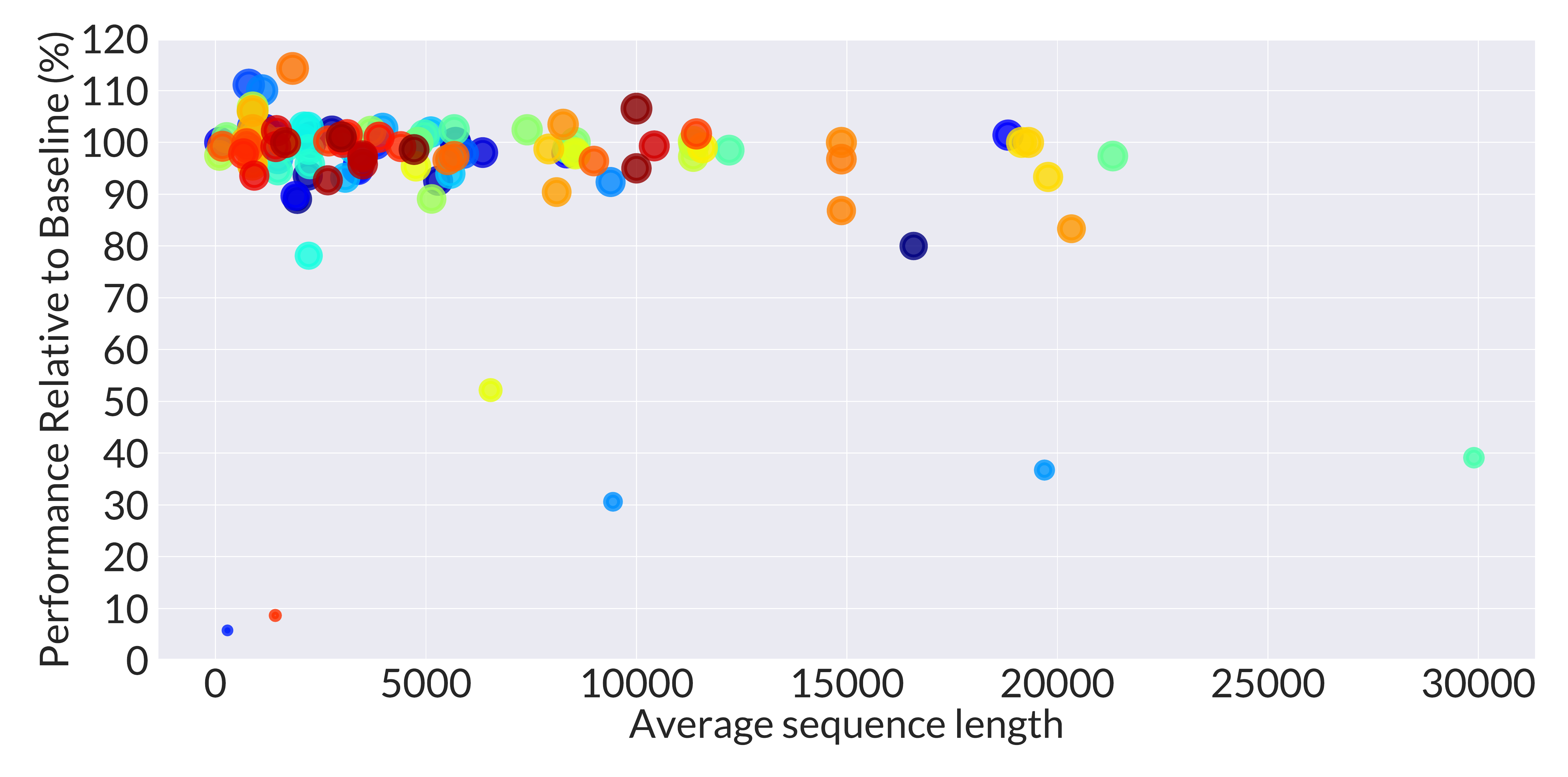}}
\subfigure[Bias: -1]
{\includegraphics[width=0.99\textwidth]{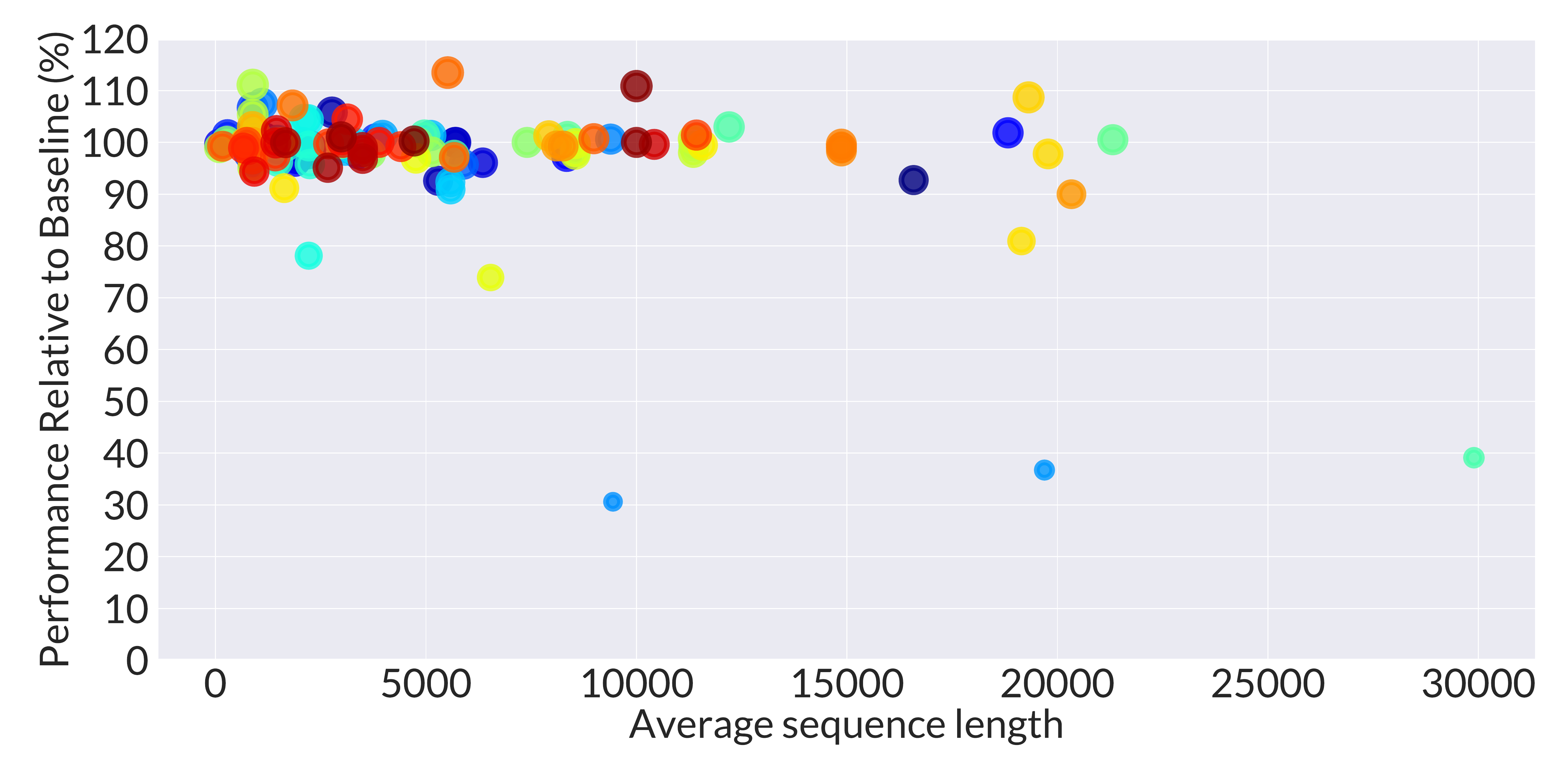}}
\end{minipage}%
\begin{minipage}{0.5\textwidth}
\captionsetup{width=.9\linewidth}
\subfigure[Bias: 1]
{\includegraphics[width=0.99\textwidth]{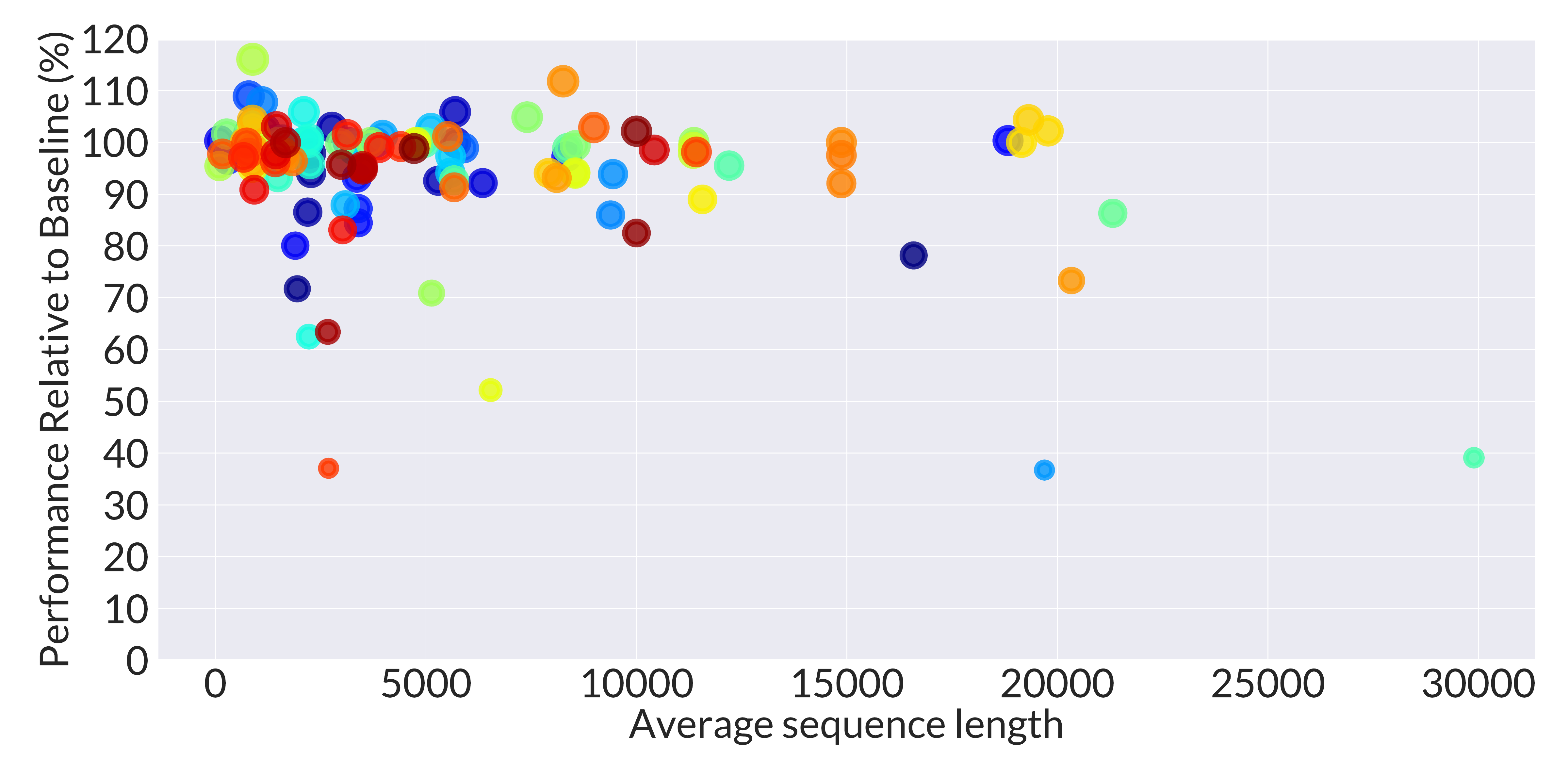}}
\caption{Improvement by an LZHRR Layer over an LSTM with different bias initializations. Points above 100\% indicate datasets on which the LZ Layer outperformed the LSTM. In (a) improvement is achieved in $44.53\%$ of the datasets, in (b) $42.19\%$ of the datasets, and in (c) $35.94\%$ of the datasets. Overall the performance in $46.88\%$ of the datasets is improved by the LZLayer. The size of the circle is proportional to the amount of improvement.}
\label{fig:improve_bias}
\end{minipage}
\end{figure*}
}
\paragraph{PS-MNIST}

\begin{wrapfigure}[8]{r}{0.45\textwidth}
\vspace{-60pt}
\centerline{\includegraphics[width=0.44\textwidth]{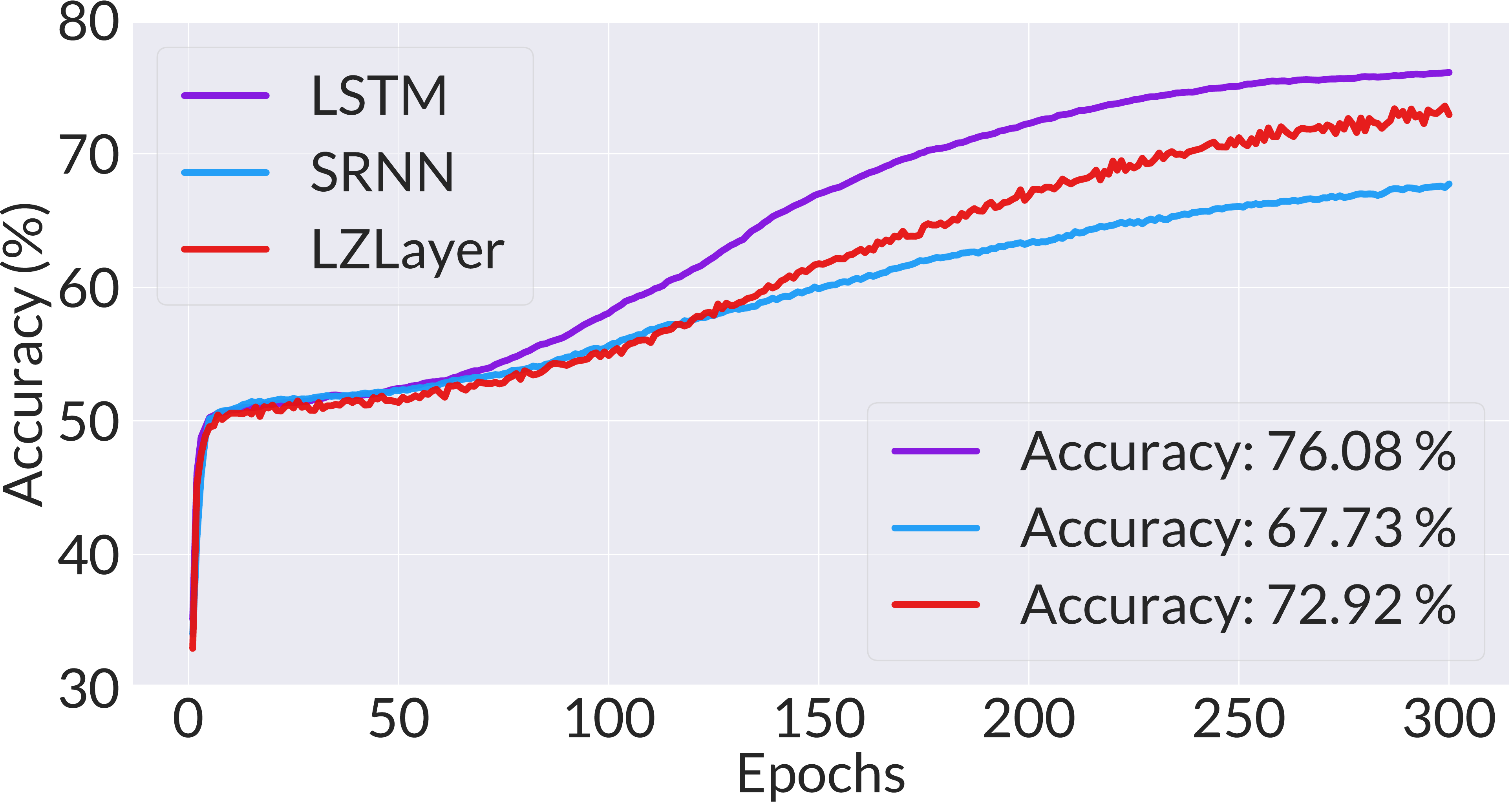}}
\caption{Accuracy curve of the LZLayer, LSTM, and SRNN on the test set of PS-MNIST.} 
\label{fig:ps-mnist}
\end{wrapfigure}

\vspace{-7pt} Permuted Sequential MNIST (PS-MNIST) \cite{le2015simple} is another benchmark for comparing recurrent networks, in which the $784$ pixels of the MNIST images are presented sequentially to an RNN following the application of a fixed permutation.  We tested the LZHRR layer on the PS-MNIST dataset and compared its performance to that of an LSTM and the Shuffle RNN (SRNN) from \cite{rotman2020shuffling}. All the recurrent networks were trained with a hidden size of 256 and a batch size of 64. The models were optimized using the Adam optimizer with a learning rate of $10^{-4}$ for 300 epochs. Figure \ref{fig:ps-mnist} shows the accuracy curve on the test set of all three models. The performance of LZLayer on PS-MNIST is better than SRNN and close to LSTM. 
 
\section{Conclusions} \label{sec:conclusion}
\vspace{-5pt} This paper proposes a novel RNN for processing long sequences, where we pair a traditional LSTM cell with an associative memory to mimic the process of LZ compression. We experimented with Hopfield Network, VTB, and HRR associative memories and ultimately settled on HRR as providing the best combination of stability, speed, and memory usage. After running our LZHRR layer on all 128 datasets of the UCR Time Series Archive, we found that the LZ network generally underperforms a traditional LSTM, regardless of sequence length. We hypothesize that this is because the HRR process is too noisy to replicate the function of the external memory in LZ compression faithfully. However, more robust associative memories come with an additional set of runtime and memory challenges, as hinted at by our experiments with Hopfield associative memories. 
\par
To improve the speed of the LZ Layer, we transitioned its code from PyTorch to JAX. To facilitate a fair comparison, we did the same for the LSTM baselines, which had been developed in PyTorch by other researchers in this area. We were surprised to find that our re-implementation improved both the runtime and the accuracy of the original baseline models. While our baselines where selected from respectable, published work \cite{rotman2020shuffling}, our results highlight that using baselines naively from prior works can lead our evaluations and conclusions astray. We hypothesize that this may be in part due to the niceness of the research area, as fewer people have spent time tuning these baselines to ensure robust performance. This mimics problems recently seen in other areas of ML reproducibility, where we have observed that improving the baselines mitigates any benefits of the new technique, resulting in ``false discoveries'' \cite{Musgrave2020,mlsys2020_73,Dacrema2019,Bouthillier2021} and increasing the effort required to replicate results \cite{Raff2020c,Raff2022a}.

\bibliography{refs}
\newpage
\appendix

\section{Supplemental Results} \label{sec:appendix}
\subsection{Addition problem} \label{sec:appendix_addprob}
\begin{figure}[H]
\centering
\includegraphics[width=0.49\textwidth]{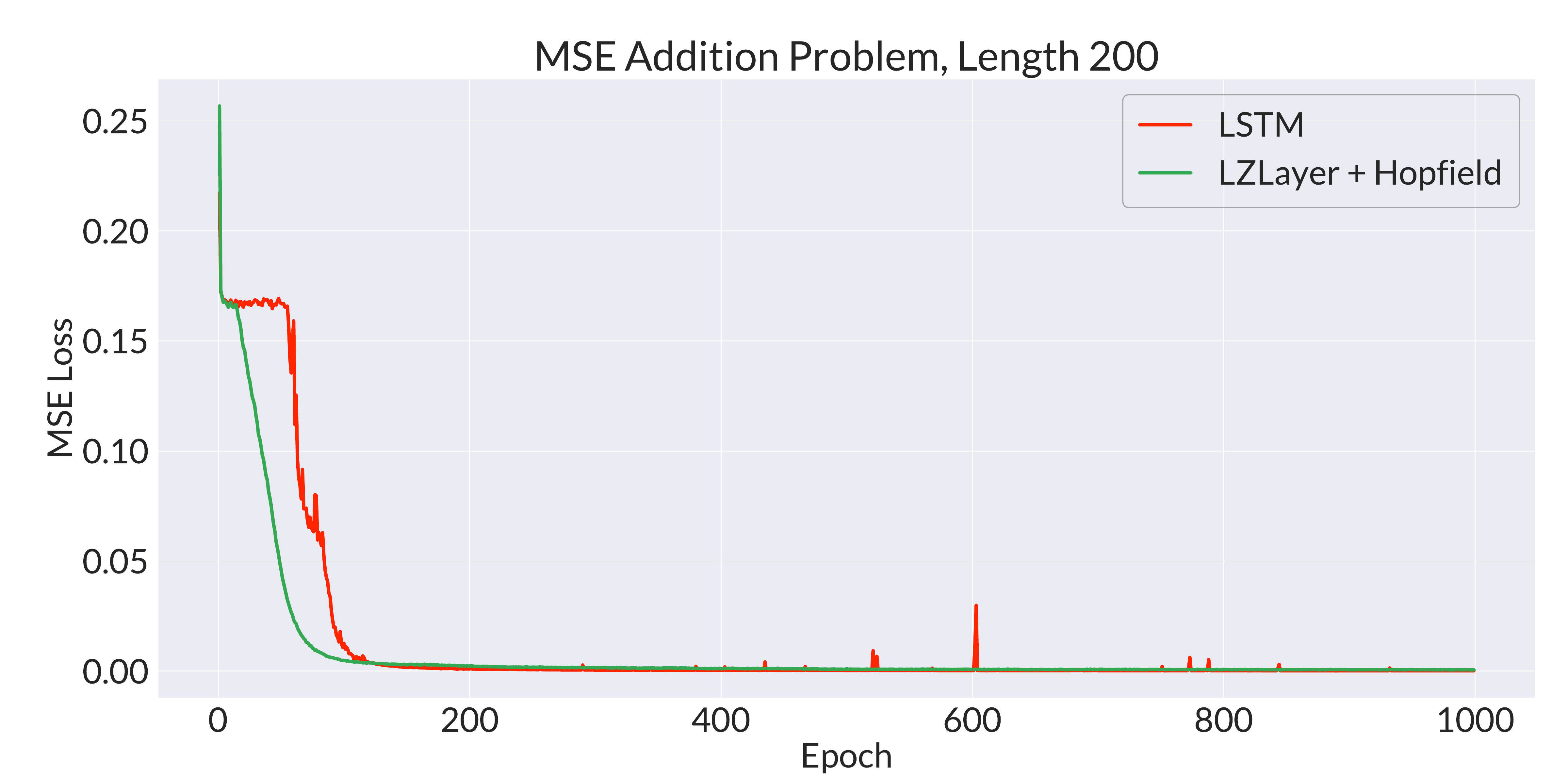}
\includegraphics[width=0.49\textwidth]{figs/MSE_addition_400_nohrr.pdf}
\caption{Addition problem results --- LZ Layer with Hopfield Associative Memory.}
\label{fig: full_lzhopresults}
\end{figure}

\begin{figure}[h]
\centering
\includegraphics[width=.49\linewidth]{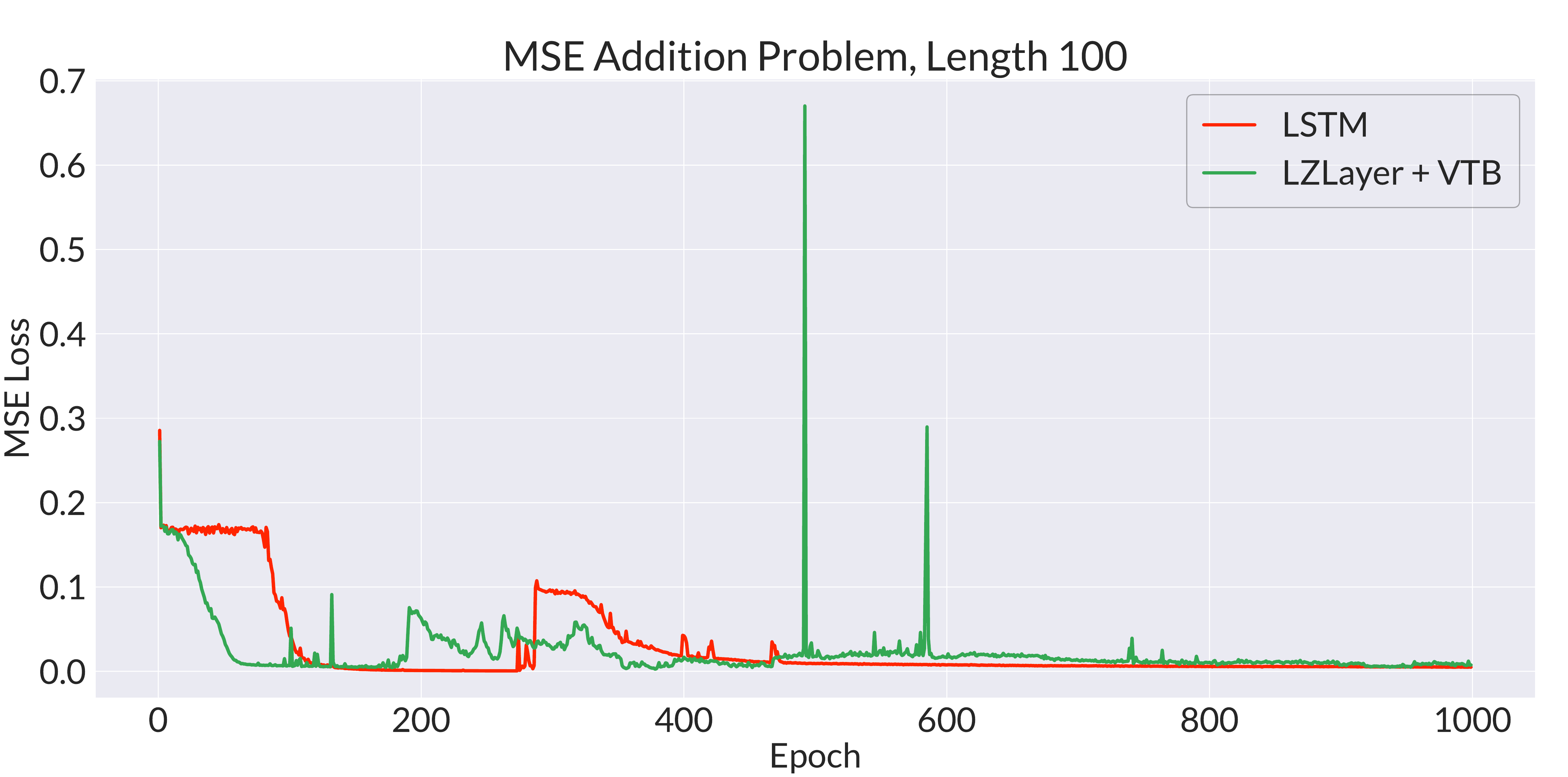}
\includegraphics[width=.49\linewidth]{figs/MSE_addition_200_nohop.pdf}
\caption{Addition problem results --- LZ Layer with VTB Associative Memory.}
\label{fig: full_lzvtbresults}
\end{figure}

\begin{figure}[h]
\centering
\includegraphics[width=.49\linewidth]{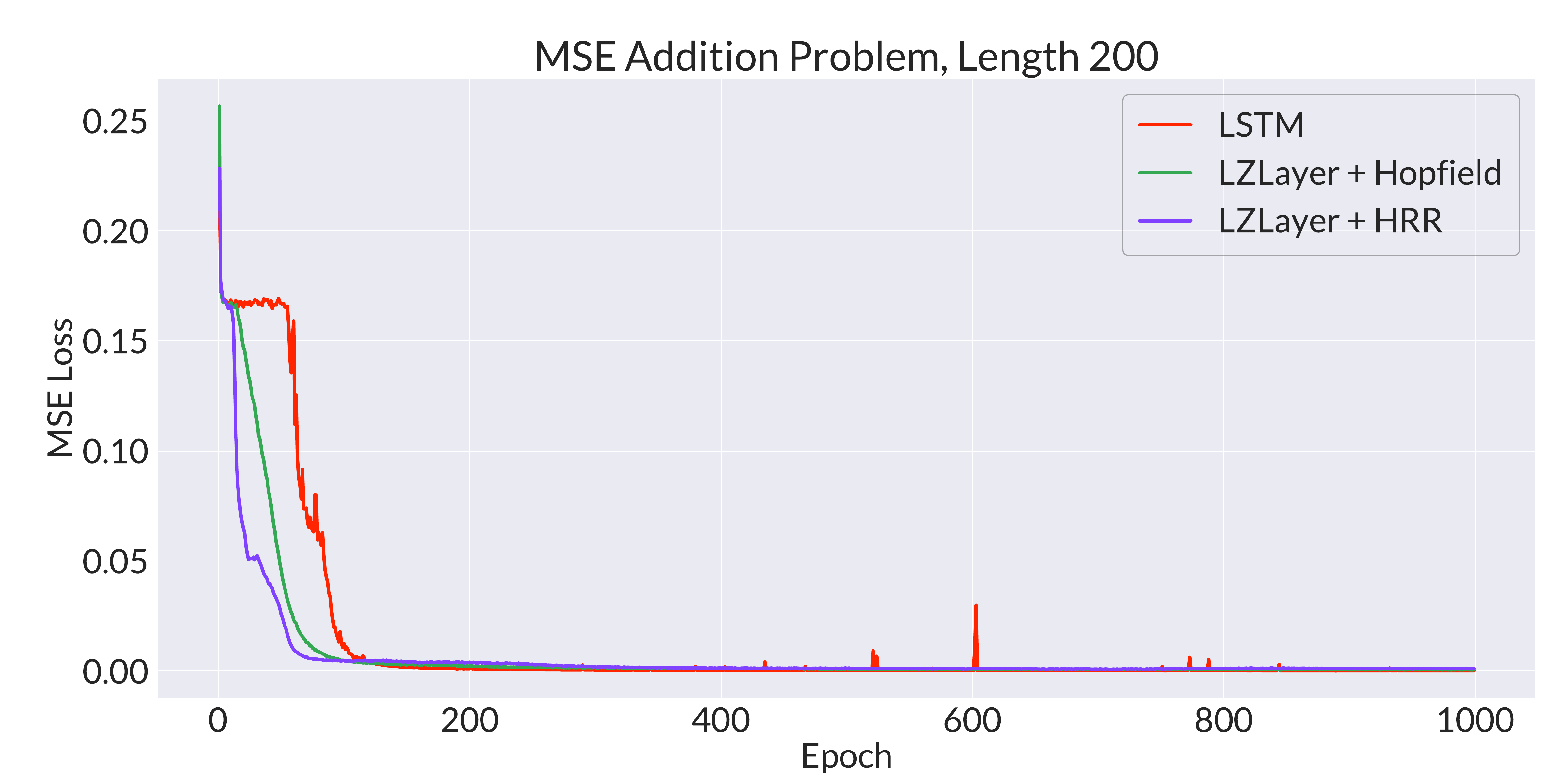}
\includegraphics[width=.49\linewidth]{figs/MSE_addition_400.pdf}
\caption{Addition problem results --- LZ Layer with HRR Associative Memory.}
\label{fig: full_lzhrrresults}
\end{figure}

\newpage
\subsection{UCR Time Series Archive} \label{sec:appendix_ucr}

\begin{table*}[h!]
\centering
\caption{UCR Dataset}
\label{tab:ucr}
\renewcommand{\arraystretch}{1.15}
\resizebox{\textwidth}{!}{%
\begin{tabular}{|c|c|c|c|c||c|c|c|c|c|}
\hline
\multirow{2}{*}{Dataset} &
  \multirow{2}{*}{LSTM} &
  \multirow{2}{*}{\shortstack{LZLayer \\$b=0$}} &
  \multirow{2}{*}{\shortstack{LZLayer \\$b=1$}} &
  \multirow{2}{*}{\shortstack{LZLayer \\$b=-1$}} &
  \multirow{2}{*}{Dataset} &
  \multirow{2}{*}{LSTM} &
  \multirow{2}{*}{\shortstack{LZLayer \\$b=0$}} &
  \multirow{2}{*}{\shortstack{LZLayer \\$b=1$}} &
  \multirow{2}{*}{\shortstack{LZLayer \\$b=-1$}} \\  &
   & & & & & & & & \\ \hline

ACSF1 & \textbf{55.0} & 44.0 & 43.0 & 51.0 & Adiac & \textbf{72.38} & 64.45 & 51.92 & 70.08 \\ \hline
AllGestureWiimoteX & 41.0 & \textbf{41.43} & 40.14 & 40.71 & AllGestureWiimoteY & \textbf{53.29} & 53.29 & 50.14 & 52.57 \\ \hline
AllGestureWiimoteZ & \textbf{44.57} & 41.71 & 38.57 & 43.57 & ArrowHead & 78.86 & 80.57 & 81.14 & \textbf{83.43} \\ \hline
Beef & \textbf{90.0} & 83.33 & 83.33 & 83.33 & BeetleFly & 85.0 & 85.0 & \textbf{90.0} & 85.0 \\ \hline
BirdChicken & \textbf{70.0} & 70.0 & 70.0 & 70.0 & BME & 79.33 & \textbf{81.33} & 81.33 & 80.0 \\ \hline
Car & \textbf{85.0} & 83.33 & 78.33 & 81.67 & CBF & 87.44 & \textbf{88.44} & 87.89 & 87.78 \\ \hline
Chinatown & 97.38 & 97.38 & \textbf{97.67} & 97.08 & ChlorineConcentration & \textbf{83.33} & 74.74 & 66.72 & 80.26 \\ \hline
CinCECGTorso & 81.96 & 83.12 & 82.25 & \textbf{83.48} & Coffee & \textbf{100.0} & 100.0 & 100.0 & 100.0 \\ \hline
Computers & \textbf{58.4} & 57.2 & 56.8 & 56.8 & CricketX & \textbf{56.15} & 53.85 & 47.44 & 55.38 \\ \hline
CricketY & \textbf{60.26} & 57.18 & 56.15 & 59.23 & CricketZ & \textbf{58.21} & 55.13 & 50.77 & 56.92 \\ \hline
Crop & 72.1 & 4.17 & 69.71 & \textbf{73.16} & DiatomSizeReduction & 95.75 & 95.42 & 95.75 & \textbf{96.41} \\ \hline
DistalPhalanxOutlineAgeGroup & \textbf{73.38} & 70.5 & 71.94 & 70.5 & DistalPhalanxOutlineCorrect & 71.38 & 73.55 & \textbf{73.91} & 72.83 \\ \hline
DistalPhalanxTW & 64.75 & 65.47 & 66.19 & \textbf{69.06} & DodgerLoopDay & 56.25 & \textbf{62.5} & 61.25 & 55.0 \\ \hline
DodgerLoopGame & 83.33 & 84.06 & \textbf{84.78} & 84.78 & DodgerLoopWeekend & \textbf{98.55} & 98.55 & 98.55 & 97.1 \\ \hline
Earthquakes & \textbf{67.63} & 66.19 & 66.91 & 64.75 & ECG200 & \textbf{90.0} & 90.0 & 89.0 & 90.0 \\ \hline
ECG5000 & \textbf{93.2} & 93.16 & 92.82 & 93.13 & ECGFiveDays & 96.05 & 96.05 & \textbf{96.52} & 96.17 \\ \hline
ElectricDevices & 49.7 & \textbf{54.69} & 53.56 & 53.4 & EOGHorizontalSignal & 43.37 & 40.06 & 37.29 & \textbf{43.65} \\ \hline
EOGVerticalSignal & \textbf{27.07} & 8.29 & 25.41 & 8.29 & EthanolLevel & \textbf{68.6} & 25.2 & 25.2 & 25.2 \\ \hline
FaceAll & \textbf{81.83} & 79.23 & 78.82 & 80.83 & FaceFour & 84.09 & \textbf{86.36} & 85.23 & 85.23 \\ \hline
FacesUCR & 79.85 & 79.37 & 79.07 & \textbf{80.2} & FiftyWords & \textbf{70.99} & 66.15 & 62.42 & 69.89 \\ \hline
Fish & 85.14 & 86.86 & \textbf{87.43} & 86.29 & FordA & \textbf{78.94} & 74.17 & 74.32 & 72.88 \\ \hline
FordB & \textbf{70.62} & 68.4 & 68.77 & 64.2 & FreezerRegularTrain & \textbf{93.19} & 93.16 & 92.98 & 93.16 \\ \hline
FreezerSmallTrain & \textbf{69.58} & 68.74 & 68.6 & 68.67 & Fungi & 85.48 & \textbf{86.02} & 85.48 & 85.48 \\ \hline
GestureMidAirD1 & 53.08 & 54.62 & 53.08 & \textbf{55.38} & GestureMidAirD2 & 52.31 & 53.85 & \textbf{55.38} & 54.62 \\ \hline
GestureMidAirD3 & \textbf{24.62} & 19.23 & 15.38 & 19.23 & GesturePebbleZ1 & 85.47 & 84.3 & \textbf{86.05} & 84.88 \\ \hline
GesturePebbleZ2 & \textbf{75.95} & 72.78 & 72.78 & 72.78 & GunPoint & \textbf{94.67} & 94.0 & 94.0 & 94.0 \\ \hline
GunPointAgeSpan & \textbf{94.3} & 89.24 & 87.97 & 90.82 & GunPointMaleVersusFemale & \textbf{99.68} & 99.37 & 99.05 & 99.05 \\ \hline
GunPointOldVersusYoung & \textbf{100.0} & 100.0 & 100.0 & 100.0 & Ham & 66.67 & \textbf{67.62} & 66.67 & 67.62 \\ \hline
HandOutlines & \textbf{91.89} & 35.95 & 35.95 & 35.95 & Haptics & 43.51 & 42.86 & 41.56 & \textbf{44.81} \\ \hline
Herring & 64.06 & \textbf{65.62} & 59.38 & 62.5 & HouseTwenty & 73.11 & 72.27 & 72.27 & \textbf{73.95} \\ \hline
InlineSkate & 34.55 & 33.64 & 29.82 & \textbf{34.73} & InsectEPGRegularTrain & \textbf{100.0} & 100.0 & 100.0 & 100.0 \\ \hline
InsectEPGSmallTrain & \textbf{100.0} & 100.0 & 99.6 & 100.0 & InsectWingbeatSound & 63.08 & \textbf{63.43} & 62.93 & 62.93 \\ \hline
ItalyPowerDemand & 95.53 & 96.4 & \textbf{97.08} & 95.72 & LargeKitchenAppliances & \textbf{43.2} & 43.2 & 42.93 & 42.4 \\ \hline
Lightning2 & 67.21 & 68.85 & \textbf{70.49} & 67.21 & Lightning7 & 64.38 & \textbf{65.75} & 64.38 & 63.01 \\ \hline
Mallat & \textbf{92.71} & 92.2 & 92.71 & 92.03 & Meat & \textbf{91.67} & 81.67 & 65.0 & 90.0 \\ \hline
MedicalImages & 70.66 & \textbf{71.32} & 68.55 & 70.39 & MelbournePedestrian & \textbf{92.78} & 90.45 & 88.56 & 91.84 \\ \hline
MiddlePhalanxOutlineAgeGroup & 52.6 & 55.84 & \textbf{61.04} & 58.44 & MiddlePhalanxOutlineCorrect & 76.29 & \textbf{81.44} & 78.69 & 80.41 \\ \hline
MiddlePhalanxTW & 53.25 & \textbf{56.49} & 54.55 & 50.65 & MixedShapesRegularTrain & \textbf{90.68} & 88.21 & 88.74 & 88.91 \\ \hline
MixedShapesSmallTrain & 84.0 & 84.16 & 83.22 & \textbf{84.54} & MoteStrain & 85.46 & 86.42 & 86.9 & \textbf{87.06} \\ \hline
NonInvasiveFetalECGThorax1 & \textbf{92.77} & 90.94 & 87.18 & 90.64 & NonInvasiveFetalECGThorax2 & \textbf{92.72} & 90.64 & 87.33 & 92.52 \\ \hline
OliveOil & \textbf{76.67} & 40.0 & 40.0 & 56.67 & OSULeaf & \textbf{53.72} & 51.24 & 53.72 & 52.07 \\ \hline
PhalangesOutlinesCorrect & \textbf{79.14} & 75.99 & 75.29 & 78.32 & Phoneme & \textbf{10.07} & 9.97 & 8.97 & 10.02 \\ \hline
PickupGestureWiimoteZ & \textbf{68.0} & 68.0 & 66.0 & 62.0 & PigAirwayPressure & \textbf{10.1} & 10.1 & 10.1 & 8.17 \\ \hline
PigArtPressure & 21.63 & 20.19 & \textbf{22.12} & 21.15 & PigCVP & 11.06 & 11.06 & 11.54 & \textbf{12.02} \\ \hline
PLAID & 44.13 & 43.58 & 41.53 & \textbf{44.69} & Plane & \textbf{98.1} & 98.1 & 96.19 & 98.1 \\ \hline
PowerCons & \textbf{100.0} & 100.0 & 100.0 & 100.0 & ProximalPhalanxOutlineAgeGroup & 80.49 & \textbf{85.37} & 83.9 & 82.93 \\ \hline
ProximalPhalanxOutlineCorrect & \textbf{85.22} & 81.44 & 81.79 & 82.47 & ProximalPhalanxTW & 80.0 & \textbf{81.95} & 79.02 & 80.0 \\ \hline
RefrigerationDevices & \textbf{38.93} & 35.2 & 36.27 & 38.67 & Rock & \textbf{60.0} & 50.0 & 44.0 & 54.0 \\ \hline
ScreenType & 38.4 & 39.73 & \textbf{42.93} & 38.13 & SemgHandGenderCh2 & \textbf{90.17} & 90.17 & 90.17 & 88.67 \\ \hline
SemgHandMovementCh2 & \textbf{67.56} & 58.67 & 62.22 & 67.33 & SemgHandSubjectCh2 & \textbf{89.33} & 86.44 & 87.11 & 88.44 \\ \hline
ShakeGestureWiimoteZ & 56.0 & \textbf{64.0} & 54.0 & 60.0 & ShapeletSim & 49.44 & 47.78 & 50.0 & \textbf{56.11} \\ \hline
ShapesAll & \textbf{75.0} & 73.0 & 68.5 & 72.83 & SmallKitchenAppliances & 37.07 & 35.73 & \textbf{38.13} & 37.33 \\ \hline
SmoothSubspace & \textbf{90.0} & 89.33 & 88.0 & 89.33 & SonyAIBORobotSurface1 & \textbf{72.55} & 71.71 & 71.05 & 71.55 \\ \hline
SonyAIBORobotSurface2 & \textbf{82.9} & 82.69 & 82.79 & 82.9 & StarLightCurves & 91.68 & \textbf{93.19} & 90.0 & 92.98 \\ \hline
Strawberry & 96.22 & \textbf{96.49} & 35.68 & 95.95 & SwedishLeaf & \textbf{84.96} & 7.36 & 81.76 & 82.72 \\ \hline
Symbols & \textbf{87.04} & 86.33 & 86.33 & 86.33 & SyntheticControl & \textbf{92.0} & 90.0 & 89.33 & 91.0 \\ \hline
ToeSegmentation1 & 58.77 & 59.65 & 59.65 & \textbf{61.4} & ToeSegmentation2 & 76.92 & \textbf{77.69} & 76.15 & 76.92 \\ \hline
Trace & \textbf{77.0} & 77.0 & 64.0 & 77.0 & TwoLeadECG & \textbf{90.43} & 84.64 & 82.18 & 85.34 \\ \hline
TwoPatterns & \textbf{92.48} & 91.73 & 90.42 & 92.38 & UMD & 92.36 & 94.44 & \textbf{95.14} & 94.44 \\ \hline
UWaveGestureLibraryAll & \textbf{95.42} & 94.78 & 94.03 & 95.06 & UWaveGestureLibraryX & \textbf{76.35} & 74.4 & 72.84 & 74.96 \\ \hline
UWaveGestureLibraryY & \textbf{69.1} & 67.11 & 65.69 & 68.43 & UWaveGestureLibraryZ & \textbf{69.71} & 66.78 & 66.08 & 67.53 \\ \hline
Wafer & \textbf{99.61} & 99.56 & 99.56 & 99.56 & Wine & \textbf{75.93} & 70.37 & 48.15 & 72.22 \\ \hline
WordSynonyms & 58.46 & \textbf{59.09} & 55.96 & 59.09 & Worms & \textbf{51.95} & 49.35 & 42.86 & 51.95 \\ \hline
WormsTwoClass & 59.74 & 63.64 & 61.04 & \textbf{66.23} & Yoga & 84.87 & 83.7 & 83.87 & \textbf{85.1} \\ \hline

\end{tabular}%
}
\end{table*}

\newpage
\subsection{JAX v PyTorch} \label{sec:appendix_jax}

\begin{figure}[H]
\vspace{-15pt}
\centering 
\subfigure[LSTM]
{\includegraphics[width=0.49\textwidth]{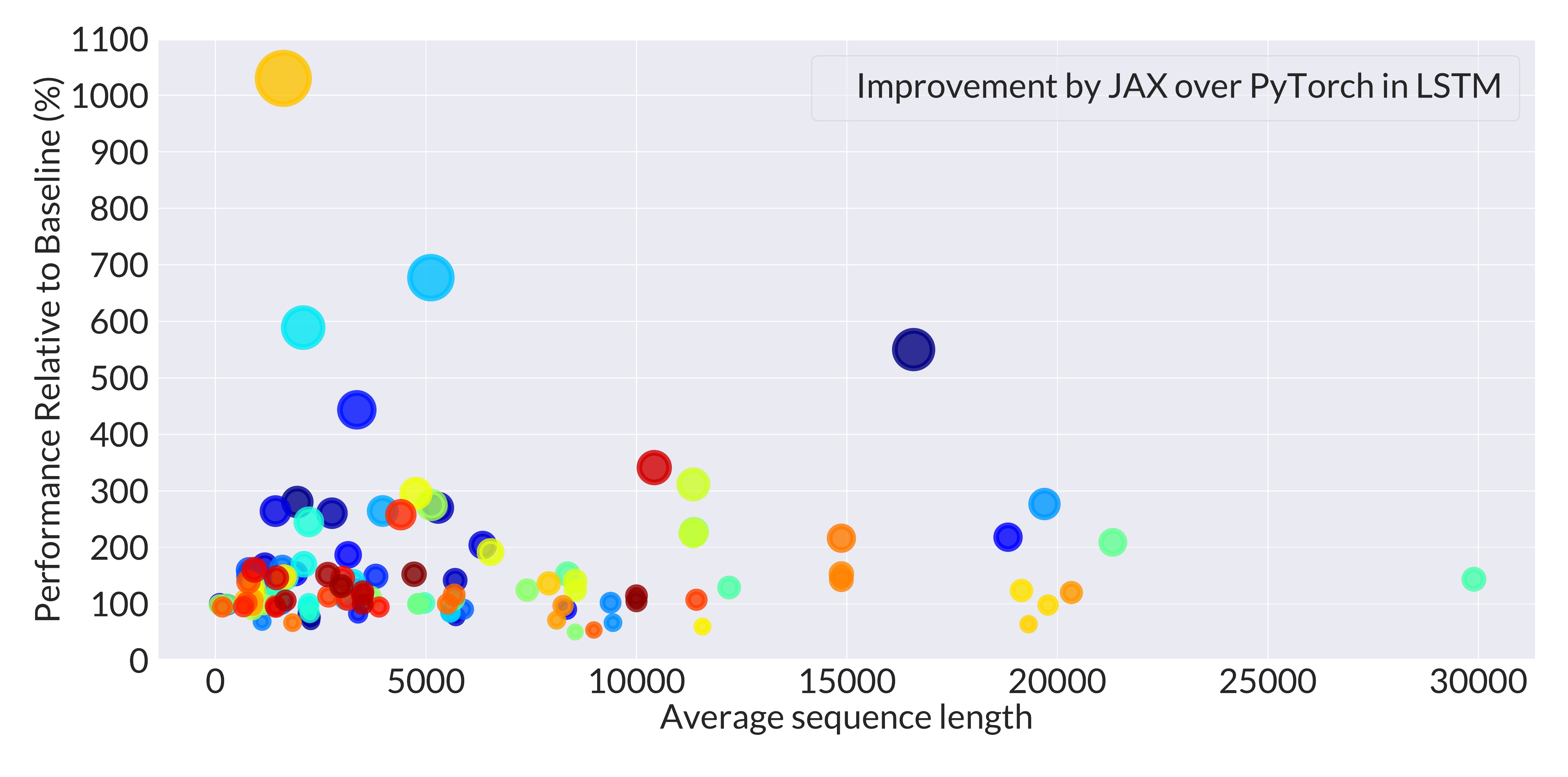}}
\subfigure[LZLayer]
{\includegraphics[width=0.49\textwidth]{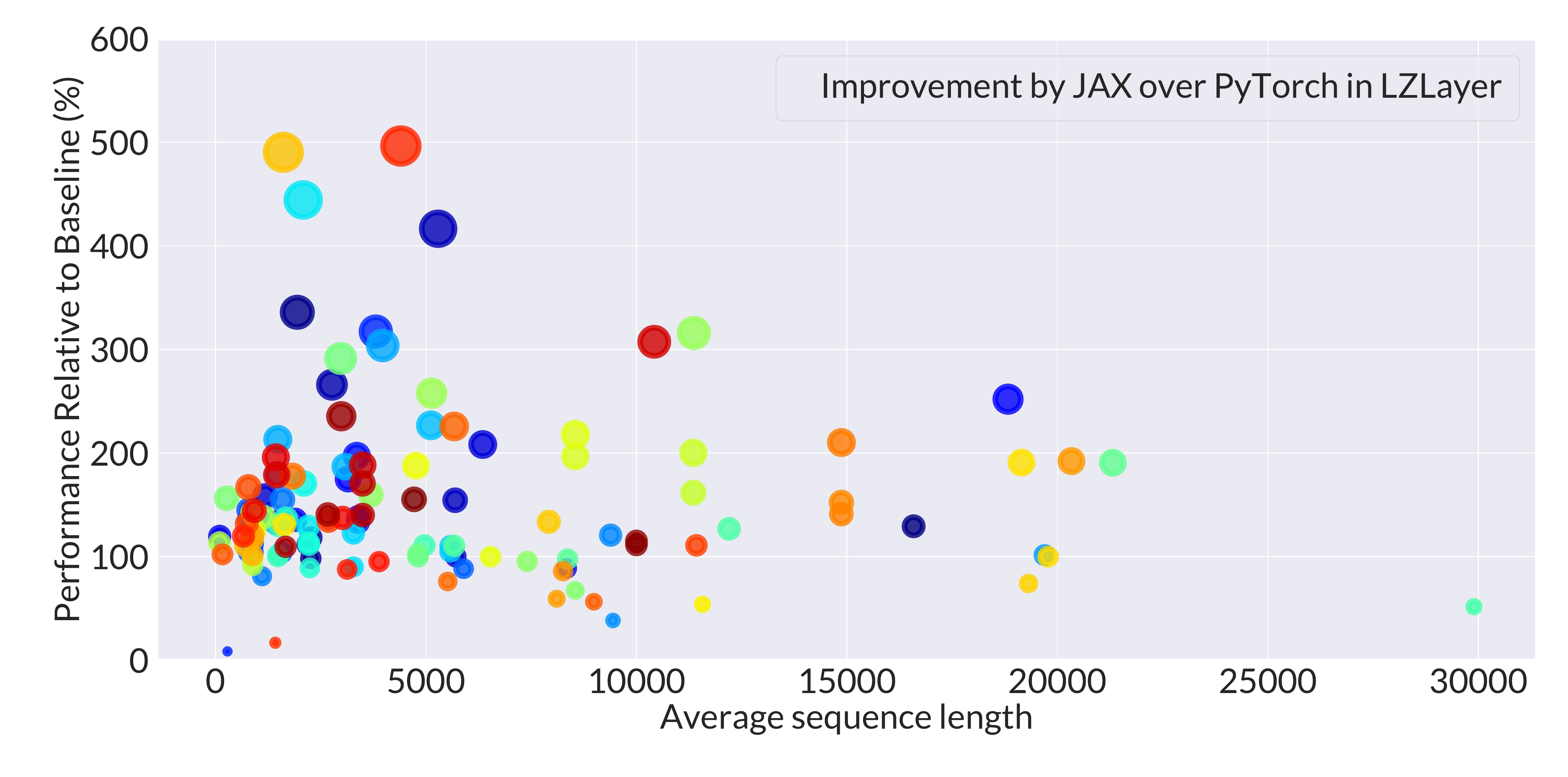}}
\caption{\small Accuracy improvement from JAX over PyTorch implementation in both LSTM (a) and LZLayer (b). Points above 100\% indicate datasets on which the implementation in JAX outperformed the implementation in PyTorch. The accuracy of the LSTM is improved in $69.53\%$ of the datasets and the accuracy of the LZ Layer is improved in $79.68\%$ of the datasets. The size of the circle is proportional to the amount of improvement.}
\label{fig:improve_jax}
\end{figure}

\subsection{Memory copy problem}
The LZHRR layer, implemented in PyTorch, was also tested on the memory copy problem. Though the results were not particularly revealing, we include them here in the appendix for completeness.

The copy problem tests an RNN's ability to memorize the first $N$ items in a long sequence. As input, the RNN receives a sequence of $N$ significant items, drawn from a fixed set $M$ of symbols, followed by $T-1$ "blank" inputs, a special delimiter symbol, and $N$ more blank items. The desired output is $N+T$ blank symbols, followed by the $N$ original significant items (see Table \ref{tab: copyprob}). We compare to a baseline model which outputs a constant sequence after it sees the delimiter symbol; this base solution has a cross-entropy of $\frac{N\ln M}{T+2N}$. 

\par
In conjunction with \cite{rotman2020shuffling}, we chose parameters of $N=10$, $M=8$, a batch size of 20, and a hidden size of 128. We tested $T=100, 200, 300, 500, 1000,$ and $2000$, training each model for 1000 epochs. Again, we use the RMSProp optimizer with a learning rate of $10^{-3}$ and a decay rate of 0.9, and a model with only one LZ Layer. We also include results from a traditional LSTM, trained with the same parameters, for an additional point of comparison.

\par 
For every choice of $T$, the LSTM outperformed the LZ Layer. However, as the sequence length increased, the gap between the performance of the LZ Layer and that of the LSTM shrunk. Interestingly, this was not due to LSTM performance deteriorating as the sequences lengthened but rather because the LZ Layer improved significantly.

\begin{table}[h]
\vspace{-7pt}
\centering
\caption{Sample input and output for the copy problem with $N=5$, $M=3$, $T=10$. $\star$ denotes a special delimiter symbol; in practice, it can be a digit not used in the memorization sequence.}
\vspace{10pt}
\begin{tabular}{|c|c|c|c|c|c|c|c|c|c|c|c|c|c|c|c|c|c|c|c|}
\hline
1 & 2 & 3 & 2 & 1 & 0 & 0 & 0 & 0 & 0 & 0 & 0 & 0 & 0 & $\star$ & 0 & 0 & 0 & 0 & 0 \\
\hline
\end{tabular}
\par \bigskip
\begin{tabular}{|c|c|c|c|c|c|c|c|c|c|c|c|c|c|c|c|c|c|c|c|}
\hline
0 & 0 & 0 & 0 & 0 & 0 & 0 & 0 & 0 & 0 & 0 & 0 & 0 & 0 & 0 & 1 & 2 & 3 & 2 & 1 \\
\hline
\end{tabular}
\label{tab: copyprob}
\end{table}

\begin{figure}[h]
\centering
\includegraphics[width=0.49\textwidth]{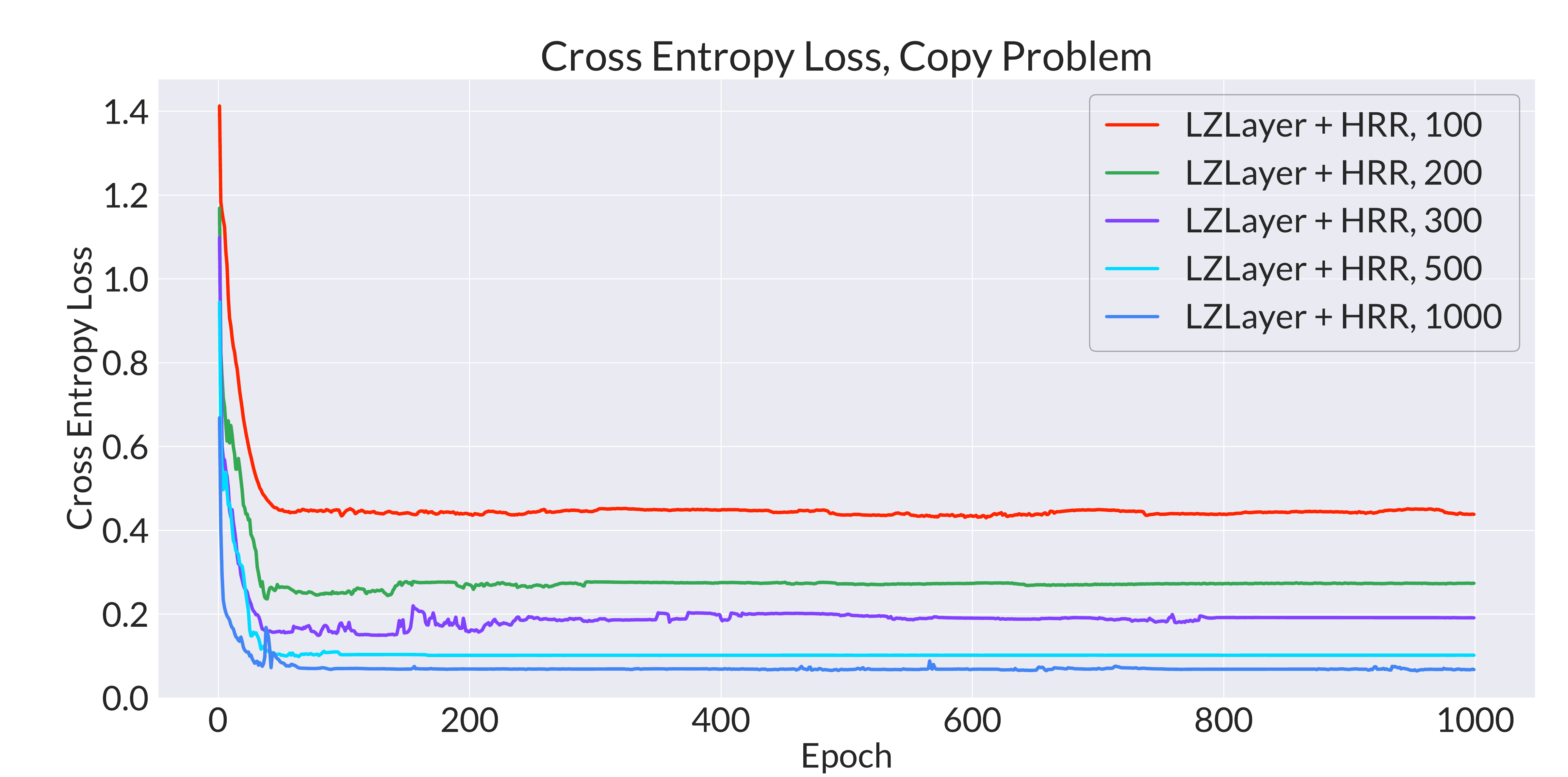}
\includegraphics[width=0.49\textwidth]{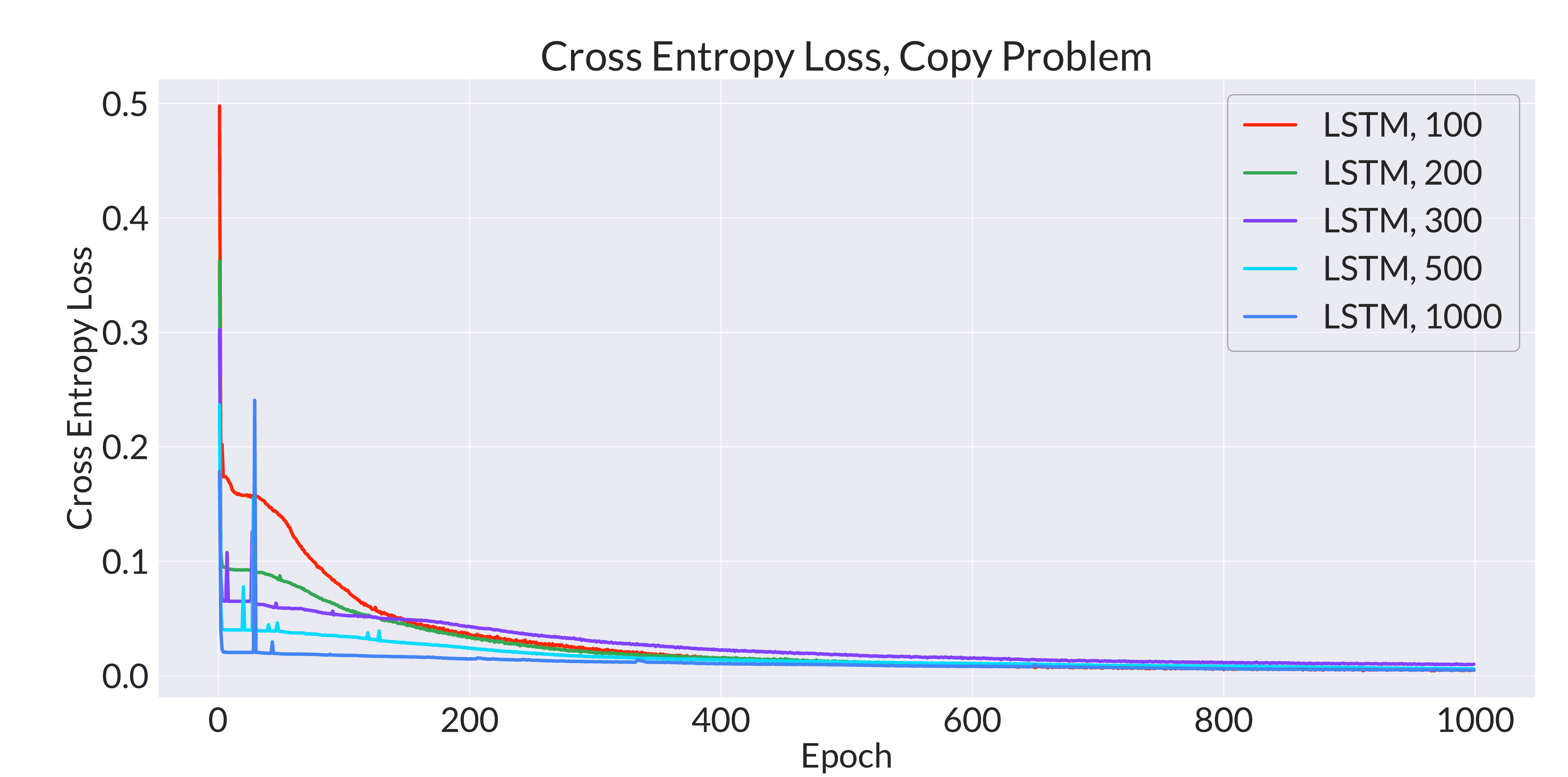}
\caption{Cross entropy losses on the memory copy problem with varying sequence lengths for an LZ Layer with HRR Associative Memory (left) and an LSTM (right).}
\label{fig:copyfig}
\end{figure}

\end{document}